# Generative Creativity:
# Adversarial Learning for Bionic Design

Simiao Yu[1], Hao Dong[1], Pan Wang[1], Chao Wu[2], and Yike Guo[1]

[1] Imperial College London

[2] Zhejiang University

**Abstract.** Bionic design refers to an approach of generative creativity in which a target object (e.g. a floor lamp) is designed to contain features of biological source objects (e.g. flowers), resulting in creative biologically-inspired design. In this work, we attempt to model the process of shape-oriented bionic design as follows: given an input image of a design target object, the model generates images that 1) maintain shape features of the input design target image, 2) contain shape features of images from the specified biological source domain, 3) are plausible and diverse. We propose DesignGAN, a novel unsupervised deep generative approach to realising bionic design. Specifically, we employ a conditional Generative Adversarial Networks architecture with several designated losses (an adversarial loss, a regression loss, a cycle loss and a latent loss) that respectively constrict our model to meet the corresponding aforementioned requirements of bionic design modelling. We perform qualitative and quantitative experiments to evaluate our method, and demonstrate that our proposed approach successfully generates creative images of bionic design.

## 1 Introduction

Generative creativity refers to the generation process of new and creative objects composing features of existing domains. In computer vision, achieving generative creativity is a long-term goal, and there exist works that involve generative creativity. For example, image style transfer [14, 22, 15] can be seen as a generative creativity process in which the creative images are generated by composing the features of existing content images and style images in a novel manner.

In this paper, we attempt to automate the process of bionic design [21, 52] by using deep generative networks. Bionic design refers to a method of product design, in which a biologically-inspired object is created by combining the features of a target design object with those of biological source objects. In this work, we mainly focus on *shape-oriented* bionic design, which is the crucial step in studying the general bionic design problem. More specifically, given an input image of the design target, we aim to generate images that 1) maintain the shape features of the input image, 2) contain the shape features of images from the biological source domain, 3) remain plausible and diverse. Fig. 1 illustrates examples of bionic design results generated by our proposed model. Essentially, bionic design is the ideal task to demonstrate generative creativity, because this process can be seen as composing the features of design target images and biological source images into novel and creative images that never existed before.

Automating the aforementioned process of bionic design is a challenging task due to the following reasons. First, the task is of unsupervised learning, since the nature of creative design implies that there is no or very few available images of biologically-inspired design. In our case, we only have unpaired data of design target images and biological source images. Second, there should be multiple ways of integrating features of biological source images into the given design target image. In other words, bionic design is a one-to-many generation process, and the learned generative model should be able to achieve this variation. Third, the generated biologically-inspired design should preserve key features of input design target image and biological source images, which requires the model to be able to select and merge the features of different sources.

We propose DesignGAN, a novel unsupervised deep generative approach for bionic design. Our method is based on the architecture of conditional generative adversarial networks (cGAN) [17, 39], with various enhancements designed to resolve the challenges mentioned above. First, the generator takes as input both an image and a latent variable sampled from a prior Gaussian distribution, which enables the model to generate diverse output images. This is implemented by the introduction of an encoder and a latent loss. Second, our approach employs both cycle loss [62, 30, 56] and regression loss to help maintain the key features of the design target. Last, an adversarial loss is used to integrate the features of biological source images into the input image.

We conduct both qualitative and quantitative experiments on the "Quick, Draw!" dataset [19], and show that our proposed model is capable of generating plausible and diverse biologically-inspired design images. Fig. 1 (c) presents examples of 3D product modelling designed by a human designer who is inspired by the generated creative images.

## 2   Related Work

**Deep generative networks.** Several deep neural network architectures for image generation have been proposed recently, such as generative adversarial networks (GAN) [17], variational autoencoders (VAE) [32, 47, 18], and autoregressive models [42, 43]. Our proposed approach is based on the GAN architecture that learns to approximate the data distribution implicitly, by training a generator and a discriminator in a competing manner. The generator of the original GAN takes as input a noise vector and can be further conditioned by taking as input other conditional information (such as labels [8], texts [46, 58] and images), which forms the conditional GAN architecture (cGAN) [39]. There are also numerous follow-up works proposed to enhance the image generation performance and training stability of GAN, in terms of new architectures [45, 12, 40, 41], objective functions [60, 3, 38] and training procedure [9, 48, 24, 29].

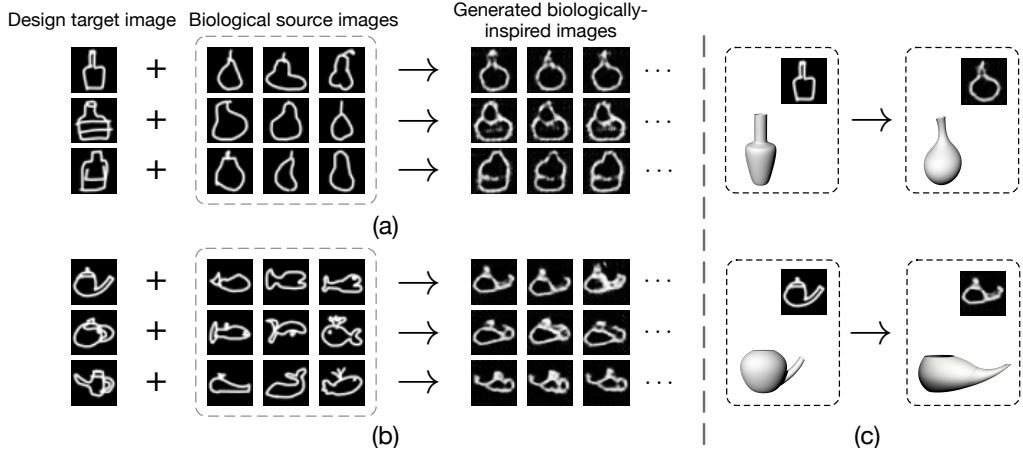

**Fig. 1.** Given a design target image and biological source images, our proposed model generates varied biologically-inspired images. (a) Wine bottle + pear, generates pear-like bottles. (b) Teapot + whale, generates whale-shaped teapots. (c) Examples of 3D product modelling designed by a human designer, inspired by the generated creative images from our proposed model (*top*: a pear-like bottle; *bottom*: a whale-shaped teapot).

**Image-to-image generation.** When conditioned on images, deep generative networks learn to solve image-to-image generation tasks, such as super resolution [33], user-controlled image editing [61, 6], image inpainting [44], colorization [59, 49], etc. Many of these tasks can be considered as a domain translation problem where the goal is to find a mapping function between source domain and target domain. This problem can be of both supervised learn-ing and unsupervised learning settings. In the supervised domain translation problem (e.g. [27, 57]), paired samples (sampled from the joint distribution of data from two domains) are observed. In the unsupervised counterpart (e.g. [36, 62, 30, 56, 35, 51, 53, 5]), only unpaired samples (sampled from the marginal distribution of data from each domain) are available. Our bionic design problem can be seen as a related task to the unsupervised image-to-image translation, as images from the design target domain and biological source domain are unpaired, and there is no existing samples of biologically-inspired images. However, a significant difference is that the generating function to be learned should be able to merge the features of images from both domains and generate biologically-inspired images that are of a third "intermediate" domain, rather than finding a mapping function between the two domains. This is detailed in the next section.

Image-to-image translation is often multi-modal: an image from source domain could be translated to multiple reasonable results (i.e. one-to-many mappings). Previous works attempt to learn this multi-modality only for supervised domain transfer problems [16, 7, 4, 55]. By contrast, our proposed approach to modelling bionic design is capable of generating diverse outputs given a single input image, in the unsupervised learning setting. Two contemporary works (AugCGAN [2] and MUNIT [25]) also attempt to model the multi-modality for unsupervised image-to-image translation by making extensions to CycleGAN [62] and UNIT [35] respectively.

**Generative creativity.** Deep generative networks for image-to-image generation have enabled the development of various creative applications in computer vision. Here "creative" means the generated results should be a novel combination of existing features (e.g. colours, textures, shapes, etc.) and did not exist in the training dataset (i.e. rather than generating a similar sample from the real data distribution). For instance, neural style transfer models [15, 28, 34, 54, 62, 13, 23] generate creative images by combining the semantic content of a given image with the style of another artwork image. Many image-to-image translation tasks involve generative creativity, such as painting-to-photo translation and object transfiguration [62]. Another work [11] synthesises novel images based on a given image and a natural language description, such that the generated images correspond to the description while maintaining other features of the given image. A recent work [50] creates innovative designs for fashion. Another recent work [37] achieves generative creativity by seamlessly copying and pasting an object into a painting. These works mainly focus on colour and texture generation or manipulation (except the work [50] that also involves the generation of shapes of abstract patterns), while the problem of bionic design in this work is mainly (semantically-related) shape-oriented, which is a more challenging task for generative creativity.

## 3   Problem Formulation

The problem of bionic design can be formulated as follows. Given a design target domain $D$ containing samples $\{d_k\}_{k=1}^{M} \in D$ (e.g. floor lamps) and a biological source domain $B$ containing samples $\{b_k\}_{k=1}^{N} \in B$ (e.g. flowers), we have the corresponding latent spaces of $D$ and $B$ (respectively $Z_d$ and $Z_b$) that contain the representations of each domain. We denote the data distribution of $D$ and $B$ as $p(d)$ and $p(b)$. We then make two key assumptions of the bionic design problem: 1) there exists an "intermediate" domain $I$ containing the generated objects of biologically-inspired design $\{\hat{i}_k\}_{k=1}^{O} \in I$, and 2) the corresponding latent space of $I$ (denoted as $Z$) contains the *merged* representations of those from $Z_d$ and $Z_b$, as illustrated in Fig. 2.

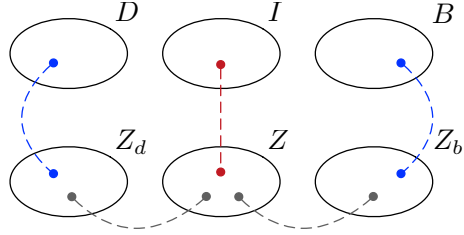

**Fig. 2.** Our assumption of the bionic design problem.

Based on these two assumptions, the objective of bionic design is to learn a generating function $G_{DB} : D \times Z \to I$, such that the generative distribution matches the distribution of $I$ (denoted as $p(i)$). Since in our case we do not have any existing samples from $I$, it is impossible to explicitly learn such generative distribution. Nonetheless, we could still learn it in an implicit fashion via real data distributions $p(d)$ and $p(b)$, and the careful design of the model architecture, as discussed in the next section. This is where generative creativity comes from. Also note that $G_{DB}$ takes as input the latent variable $z \in Z$ sampled from the distribution $p(z)$, the requirement of variations for bionic design is satisfied directly: multiple samples based on a single $d$ can then be generated by sampling different $z$ from $p(z)$.

## 4   Methodology

At first glance, the shape-oriented bionic design problem can be tackled by employing the CycleGAN architecture [62, 30, 56]. However, we reveal the significant limitations of CycleGAN for this problem, which motivates our development of a new architecture. In this section, we start with a brief discussion on the applicability and limitations of CycleGAN model and its extensions. We then describe in detail our proposed DesignGAN model and the corresponding objective functions.

## 4.1   A Path of Evolution from CycleGAN

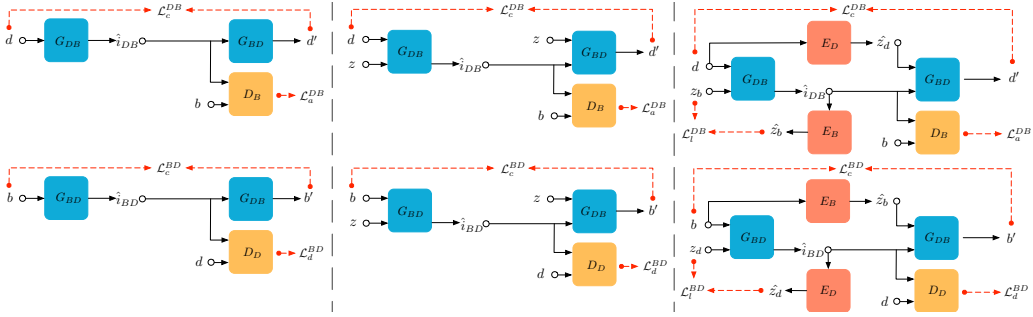

**Fig. 3.** Schema of CycleGAN model and its extensions. We explicitly present the separate components of the models to illustrate the dual learning process and the loss functions more clearly. (a) CycleGAN [62, 30, 56]. (b) CycleGAN+N. (c) CycleGan+2E.

Our initial choice is to employ the **CycleGAN** architecture directly, where two image-based cGAN models are cascaded and trained jointly (Fig. 3 (a)). We use the cycle loss to maintain the features of the given design target image and an adversarial loss to integrate the features of biological source images. Since the images only contain representations of shapes, the two losses will be forced to directly compete with each other, which makes it possible to generate images from the "intermediate" domain that contains shape features of both domains. However, this model will only learn a deterministic mapping, which will not be able to generate diverse results.

A straightforward way to make the system learn a one-to-many mapping is to inject noise as the input of the system (Fig. 3 (b)). The limitation of this approach, as also discussed in [2, 25], is that the cycle-consistence restriction would make the generator input. This is because each generator will be under conflicting constraints imposed by each cycle loss (respectively one-to-many and many-to-one mappings), which would eventually be degenerated into one-to-one mappings. We denote this model as **CycleGAN+N**.

We further propose a new architecture to solve the limitation of CycleGAN+N by integrating two encoders $E_D$, $E_B$ into the architecture (Fig. 3 (c)). Each encoder takes as input a generated image and encodes it back to the corresponding latent space. The generated latent code is used to compute a latent loss to match the input noise vector, which enforces the generator to generate diverse results. The system will never input because of this latent loss, thus resolves the problem of CycleGAN+N. However, the generated images of this system will heavily depend on the latent variable, without taking into account the input image. More specifically, given an input image $d$ and different noise vectors $z_b$, diverse $\hat{i_{DB}}$ should be generated by $G_{DB}$ because of the latent loss $\mathcal{L}_l^{DB}$. The problem emerges when calculating the cycle loss $\mathcal{L}_c^{DB}$. $G_{BD}$ is supposed to map all generated diverse images back to the original design target image $d$. Since $d$ is encoded into a fixed $\hat{z}_d$ by the encoding function $E_D$, $G_{BD}$ would simply learn a one-to-one mapping from $\hat{z}_d$ to $d$. In other words, the generators will tend to ignore the input images. We denote this model as **CycleGAN+2E**.

## 4.2 DesignGAN

To address the problem of CycleGAN+2E, we propose a new model, denoted as **DesignGAN**, as illustrated in Fig. 4. Specifically, DesignGAN is comprised of five functions parametrized by deep neural networks (two generators $G_{DB}$ and $G_{BD}$, two discriminators $D_B$ and $D_D$, and one encoder $E$) and four designated loss functions that are discussed in detail as follows. Our model is end-to-end, with all component networks trained jointly.

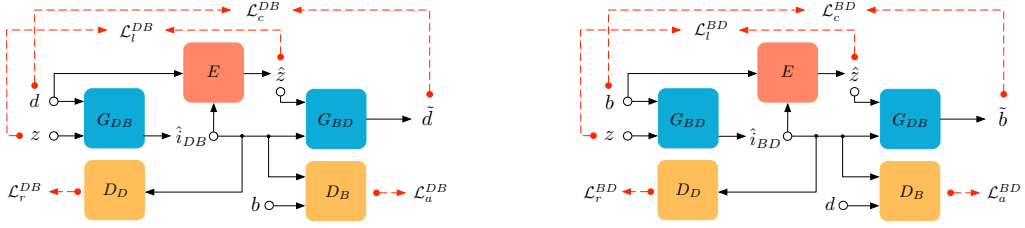

**Fig. 4.** Schema of our proposed DesignGAN model that has two key enhancements on CycleGAN+2E. First, our model employs a single encoder $E$ as the encoding function $E : B \times D \to Z$ to learn the variation of the bionic design problem. Second, we further propose to use the discriminators $D_B$ and $D_D$ simultaneously as forward regression functions to preserve the features of the input image domain, imposed by the regression loss $\mathcal{L}_r^{DB}$ and $\mathcal{L}_r^{BD}$, without competing with the generators.

**Adversarial loss.** We employ two sources of adversarial loss $\mathcal{L}_a^{DB}(G_{DB}, D_B)$ and $\mathcal{L}_a^{BD}(G_{BD}, D_D)$ that respectively enforce the outputs of $G_{DB}$ and $G_{BD}$ to match the empirical data distribution $p(b)$ and $p(d)$, as an approach to integrate corresponding features to the generated images.

$$
\begin{aligned}
\mathcal{L}_a(G_{DB}, G_{BD}, D_B, D_D) &= \mathcal{L}_a^{DB}(G_{DB}, D_B) + \mathcal{L}_a^{BD}(G_{BD}, D_D) \\
\mathcal{L}_a^{DB}(G_{DB}, D_B) &= \mathbb{E}_{b \sim p(b)}[\log D_B(b)] + \mathbb{E}_{d \sim p(d), z \sim p(z)}[\log(1 - D_B(G_{DB}(d, z)))] \\
\mathcal{L}_a^{BD}(G_{BD}, D_D) &= \mathbb{E}_{d \sim p(d)}[\log D_D(d)] + \mathbb{E}_{b \sim p(b), z \sim p(z)}[\log(1 - D_D(G_{BD}(b, z)))]
\end{aligned}
\tag{1}
$$

where $D_B$ and $D_D$ are discriminators that distinguish between generated and real images from $B$ and $D$.

**Cycle loss.** The problem of bionic design requires the generated images to maintain the features of the input design target. In other words, the generated image should still be recognised as in the class of the design target. For the shape-oriented bionic design problem, it simply implies that the generated images should resemble the input images to a large extent. After all, it would be unreasonable to generate biologically-inspired images that in turn share no relationship to the input design target image. We apply cycle loss $\mathcal{L}_c^{DB}$ and $\mathcal{L}_c^{BD}$ to constrict the generators $G_{DB}$ and $G_{BD}$ to retain the shape representations of the input images:

$$
\begin{aligned}
\mathcal{L}_c(G_{DB}, G_{BD}) &= \mathcal{L}_c^{DB}(G_{DB}, G_{BD}) + \mathcal{L}_c^{BD}(G_{BD}, G_{DB}) \\
\mathcal{L}_c^{DB}(G_{DB}, G_{BD}) &= \mathbb{E}_{d \sim p(d), z \sim p(z)}[\|G_{BD}(G_{DB}(d, z), E(G_{DB}(d, z), d)) - d\|_2^2] \\
\mathcal{L}_c^{BD}(G_{BD}, G_{DB}) &= \mathbb{E}_{b \sim p(b), z \sim p(z)}[\|G_{DB}(G_{BD}(b, z), E(b, G_{BD}(b, z))) - b\|_2^2]
\end{aligned}
\tag{2}
$$

where we employ L2 norm in the loss. The inclusion of cycle loss makes our model optimised in a dual-learning fashion [62, 30, 56]: we introduce an auxiliary generator $G_{BD}$ and train all the generators and discriminators jointly. After training, only $G_{DB}$ will be used for bionic design purpose.

**Regression loss.** The cycle loss enforces the generated images to maintain the shape features of the input image only. Another way of maintaining the design target features is to simultaneously force the generated images to contain key features of the design target domain, which directly makes the generated images recognised as the class of the design target. We therefore introduce the regression loss $L_r^{DB}$ and $L_r^{BD}$ imposed by the discriminator $D_D$ and $D_B$. $L_r^{DB}$ and $L_r^{BD}$ respectively constricts $G_{DB}$ and $G_{BD}$ to maintain representations from the domain of input images. Note that in such a situation $D_D$ and $D_B$ are employed as a regression function only, without competing with the generators as the adversarial loss does. This is why in Fig. 4 there is only one input to $D_D$ and $D_B$ when referring to $\mathcal{L}_r$. The regression loss if one of the major extensions of the CycleGAN architecture.

$$\mathcal{L}_r(G_{DB}, G_{BD}) = \mathcal{L}_r^{DB}(G_{DB}) + \mathcal{L}_r^{BD}(G_{BD})$$
$$\mathcal{L}_r^{DB}(G_{DB}) = \mathbb{E}_{d \sim p(d), z \sim p(z)}[\log(1 - D_D(G_{DB}(d, z)))] \tag{3}$$
$$\mathcal{L}_r^{BD}(G_{BD}) = \mathbb{E}_{b \sim p(b), z \sim p(z)}[\log(1 - D_B(G_{BD}(b, z)))]$$

**Latent loss.** We employ a unified encoder $E$ and a latent loss to model the variation of the bionic design problem:

$$\mathcal{L}_l(G_{DB}, G_{BD}, E) = \mathcal{L}_l^{DB}(G_{DB}, E) + \mathcal{L}_l^{BD}(G_{BD}, E)$$
$$\mathcal{L}_l^{DB}(G_{DB}, E) = \mathbb{E}_{d \sim p(d), z \sim p(z)}[\|E(G_{DB}(d, z), d) - z\|_1] \tag{4}$$
$$\mathcal{L}_l^{BD}(G_{BD}, E) = \mathbb{E}_{b \sim p(b), z \sim p(z)}[\|E(b, G_{BD}(b, z)) - z\|_1]$$

Unlike the encoders of CycleGAN+2E that take as input one image, the encoder $E$ of DesignGAN encodes a pair of images from each domain (either $(\hat{i_{DB}}, d)$ or $(b, \hat{i_{BD}})$) into the latent space $Z$ of domain $I$, which acts as an encoding function $E : B \times D \rightarrow Z$ and corresponds to our assumption of the bionic design problem. The latent loss is computed by the L1 norm distance between the generated latent variable $\hat{z}$ and the input noise vector $z$, which forces the model to generate diverse output images. More importantly, this choice of encoder ensures that neither the generated images nor the generative latent variable will be ignored under the cycle consistent constraints. This is another major extension to the CycleGAN architecture that addresses the limitation of both CycleGAN+N and CycleGAN+2E model.

**Full objective.** The full objective function of our model is:

$$\min_{\{G_{DB}, G_{BD}, E\}} \max_{\{D_B, D_D\}} \mathcal{L}(G_{DB}, G_{BD}, E, D_B, D_D) = \lambda_a \mathcal{L}_a(G_{DB}, G_{BD}, D_B, D_D) +$$
$$\lambda_c \mathcal{L}_c(G_{DB}, G_{BD}) + \lambda_r \mathcal{L}_r(G_{DB}, G_{BD}) + \lambda_l \mathcal{L}_l(G_{DB}, G_{BD}, E) \tag{5}$$

where we employ $\lambda_a$, $\lambda_c$, $\lambda_r$ and $\lambda_l$ to control the strength of individual loss components.

# 5 Experiments

**Methods.** All models discussed in Section 4, including CycleGAN, Cycle-GAN+N, CycleGAN+2E and DesignGAN, are evaluated. We employ the same network architecture in all models for a fair comparison.

**Dataset.** We evaluate our models on "Quick, Draw!" dataset [19] that contains millions of simple grayscale drawings of size 28×28 across 345 common objects. It is an ideal dataset for the shape-oriented bionic design problem. We select several pairs of domains of design targets and biological sources as the varied bionic design problems. We randomly choose 4000 images from each domain of the domain pairs for training.

**Network architecture.** For the generator networks, we adopt the encoder-decoder architecture. The encoder contains three convolutional layers and the decoder has two transposed convolutional layers. Six residual units [20] are applied after the encoder. The latent vector is spatially replicated and concatenated to the input image, where applicable. The discriminator networks contain four convolutional layers. For the encoder network, the two input images are concatenated and encoded by three convolutions and six residual units. We employ ReLU activation in the generators and encoder, and leaky-ReLU activation in the discriminators. Batch normalisation [26] is implemented in all networks.

**Training details.** The networks are trained for 120 epochs using Adam optimiser [31] with a learning rate of 0.0001 and a batch size of 64. The learning rate is decayed to zero linearly over the last half number of epochs. Due to the distinct complexity of images from different domains, the values of $\lambda_a$, $\lambda_c$, $\lambda_r$ and $\lambda_l$ and dimension of latent variable $z$ are set independently for each of the domain pairs. We use the objective functions of Least Squares GAN [38] to stabilise the learning process. The discriminator is updated using a history of generated images, as proposed in [51], in order to alleviate the model oscillation problem [62]. We apply random horizontal flipping and random $\pm 15$ degree rotation to the training images, which are further resized to $32{\times}32$ before being fed into the models. The implementation is in TensorFlow [1] and TensorLayer [10].

**Qualitative results.** Fig. 5 illustrates the qualitative comparison results of our investigated and proposed models. We maintain the same value of the latent variable for the corresponding three generated images for each group of generation, where possible. Specifically, CycleGAN in some cases is able to generate images of bionic design, while in other cases it fails to maintain features of the input design target image (e.g. Fig. 5 (d)). Also, since it is a deterministic model, no variation is produced. Similar to CycleGAN, in most cases CycleGAN+N only generates a single result given one input image, which indicates that the input noise vector is completely ignored. Although CycleGAN+2E can generate diverse results, they are either of low-quality (e.g. Fig. 5 (b) (c)), or failed to maintain any features of the input design target image. We observe that the input latent variable dominates and the design target image is ignored by CycleGAN+2E, which corresponds to our analysis in Section 4.1. By contrast, DesignGAN is capable of generating plausible and diverse biologically-inspired images that successfully maintain representations of both input design target image and biological source images.

**Quantitative results.** How to quantitatively evaluate the performance of generative models for creative tasks remains a challenging problem. In this work, we leverage human judgement to evaluate our investigated and proposed models for bionic design. Despite subjective factors being involved, it is the most dependable measurement of creativity and plausibility of generated results. More specifically, we use 8 domain pairs of design targets and biological sources shown in this paper. For each pair of domains, we select 10 images of design target as the input to our models. We then generate 3 output biologically-inspired images for every input image. There are 25 subjects recruited, shown all the input and generated images, and required to rank the models (from 1 to 4, 1 for the best) based on whether the generated images 1) maintain the key features of input design target image, 2) contain the key features of biological source domain, 3) are diverse, and 4) are creative and plausible.

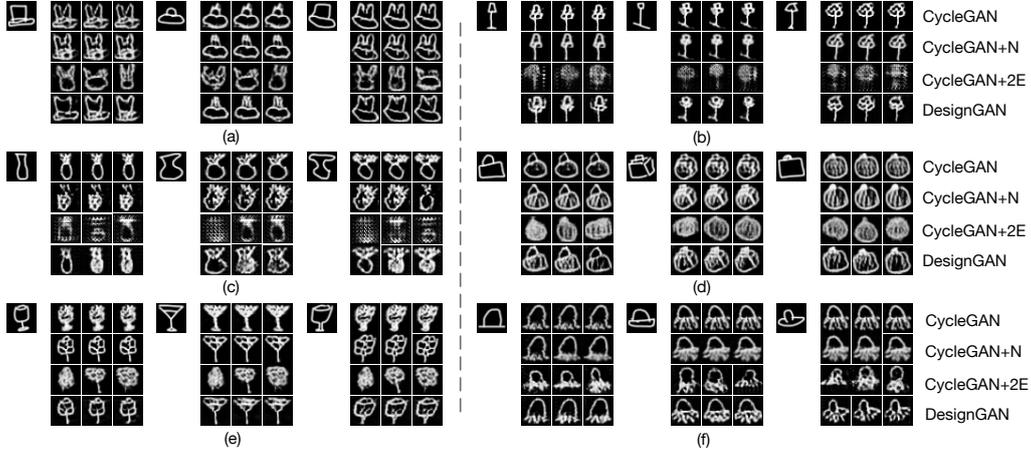

**Fig. 5.** Qualitative results of our investigated and proposed models for bionic design. (a) Hat + rabbit. (b) Floor Lamp + flower. (c) Vase + pineapple. (d) Suitcase + onion. (e) Wine glass + flower. (f) Hat + octopus.

We then average all the ranking scores and calculate the final overall scores for each model, which is presented in Table 1. All models are capable of integrating the features of biological source images to the generated images, but our proposed DesignGAN performs best in terms of maintaining the features of design target image. Although DesignGAN ranks second in diversity (CycleGAN+2E ranks first, but many of its generated images are either low-quality or failed to maintain any design target features), it gains the highest score of creativity and plausibility. Overall, DesignGAN performs best when all aspects of judging criteria are considered.

**Table 1.** Human evaluation results of our investigated and proposed models for bionic design.

|                                    | CycleGAN | CycleGAN+N | CycleGAN+2E | DesignGAN |
|------------------------------------|----------|------------|-------------|-----------|
| Maintain design target features    | 2.125    | 2.715      | 3.620       | 1.540     |
| Integrate biological source features | 2.290  | 2.690      | 2.780       | 2.240     |
| Diversity                          | 4.000    | 2.820      | 1.580       | 1.600     |
| Creativity and plausibility        | 2.455    | 2.675      | 3.545       | 1.325     |
| Overall                            | 2.718    | 2.725      | 2.881       | 1.676     |

**Comparison of regression loss and cycle loss.** We study the effect of regression loss and cycle loss by setting varied values to $\lambda_r$ and $\lambda_c$ and generating the corresponding images, which can be seen in Fig. 6. Both regression loss and cycle loss are able to improve the generated images by forcing them to contain the features of the input image (e.g. see the area pointed by the red arrow). However, if the cycle loss is applied alone, it is only when setting the weight to a relatively large value that the generated images will resemble the input image. In this case, the results tend to lose the details of features of biological source domain. By contrast, applying the regression loss makes the model generate better images, though the weight of regression loss $\lambda_r$ needs to be set to a reasonable value, in order to prevent the generated images from being exactly identical to the input image (i.e. not able to integrate the representations of the biological source domain).

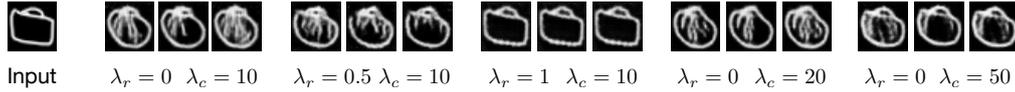

**Fig. 6.** Comparison results of regression loss and cycle loss using an example of Suitcase + onion.

**Latent variable interpolation** Fig. 7 shows the generated biologically-inspired design images by linearly interpolating the input latent variable $z$. The smooth semantic transitions of generated results verify that our model learns a smooth latent manifold as well as the disentangled representations for the bionic design problem.

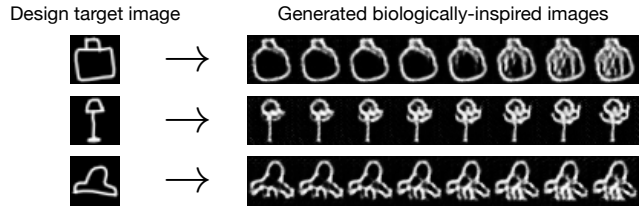

**Fig. 7.** Generated biologically-inspired design images by interpolating the input latent variable. (*top*) Suitcase + onion. (*middle*) Floor lamp + flower. (*bottom*) Hat + octopus.

## 6    Conclusions

In this paper, we proposed DesignGAN as a novel unsupervised deep generative network with the capacity of shape-oriented bionic design. We presented a systemic design path of this architecture. The research shows how the CycleGAN architecture can be further evolved into an adversarial learning framework with strong generative creativity. We conducted qualitative and quantitative experiments on the methods of our design path and demonstrated that our proposed model achieves superior results of plausible and diverse biologically-inspired design images. The shape-oriented bionic design problem we addressed can be regarded as an essential prerequisite for tackling more comprehensive and complicated bionic design problems that may require the manipulation of colours, textures, etc. Another direction of future work is to model the bionic design problem in a goal-oriented manner as human designers would (rather than random generation) where advanced technologies such as reinforcement learning can be applied.